\documentclass{article}

\usepackage[preprint]{neurips_data_2023}

\usepackage[utf8]{inputenc} 
\usepackage[T1]{fontenc}    
\usepackage{hyperref}       
\usepackage{url}            
\usepackage{booktabs}       
\usepackage{amsfonts}       
\usepackage{nicefrac}       
\usepackage{microtype}      
\usepackage{xcolor}         
\usepackage{CJKutf8}
\usepackage{multirow}
\usepackage{amssymb}
\usepackage{amsmath}
\usepackage{amsthm}
\usepackage{bbding}
\usepackage{graphicx}
\usepackage{makecell}

\title{PIXIU: A Large Language Model, Instruction Data and Evaluation Benchmark for Finance}

\author{
Qianqian Xie\\
School of Computer Science\\
Wuhan University\\
Wuhan, Hubei, China\\
\texttt{xieq@whu.edu.cn}\\
\And
Weiguang Han\\
School of Computer Science\\
Wuhan University\\
Wuhan, Hubei, China\\
\texttt{han.wei.guang@whu.edu.cn}\\
\And
Xiao Zhang\\
Sun Yat-Sen University\\
Shenzhen, Guangdong, China\\
\texttt{zhangx767@mail2.sysu.edu.cn}\\
\And
Yanzhao Lai\\
School of Economics and Management\\
Southwest Jiaotong University\\
Chengdu, Sichuan, China\\
\texttt{laiyanzhao@swjtu.edu.cn}\\
\And
Min Peng\\
School of Computer Science\\
Wuhan University\\
Wuhan, Hubei, China\\
\texttt{pengm@whu.edu.cn}\\
\And
Alejandro Lopez-Lira\\
University of Florida\\
\texttt{alejandro.lopez-lira@warrington.ufl.edu}\\
\And
Jimin Huang\\
ChanceFocus AMC.\\
Shanghai, China\\
\texttt{jimin@chancefocus.com}\\
}

\begin{document}

\maketitle

\begin{abstract}
  Although large language models (LLMs) has shown great performance on natural language processing (NLP) in the financial domain, there are no publicly available financial tailtored LLMs, instruction tuning datasets, and evaluation benchmarks, which is critical for continually pushing forward the open-source development of financial artificial intelligence (AI).
  This paper introduces PIXIU, a comprehensive framework including the first financial LLM based on fine-tuning LLaMA with instruction data, the first instruction data with 136K data samples to support the fine-tuning, and an evaluation benchmark with 5 tasks and 9 datasets.
  We first construct the large-scale multi-task instruction data considering a variety of financial tasks, financial document types, and financial data modalities.
  We then propose a financial LLM called FinMA by fine-tuning LLaMA with the constructed dataset to be able to follow instructions for various financial tasks.
  To support the evaluation of financial LLMs, we propose a standardized benchmark that covers a set of critical financial tasks, including five financial NLP tasks and one financial prediction task. 
With this benchmark, we conduct a detailed analysis of FinMA and several existing LLMs, uncovering their strengths and weaknesses in handling critical financial tasks. 
The model, datasets, benchmark, and experimental results are open-sourced~\footnote{\url{https://github.com/chancefocus/PIXIU}} to facilitate future research in financial AI.
\end{abstract}

\section{Introduction}
\label{sec:introduction}
Financial technology (FinTech) has been continually advanced by the development of natural language processing (NLP) and machine learning (ML) techniques, unlocking diversity capabilities from predicting stock price movements to advanced financial analytics~\citep{araci2019finbert,han2023select,xie2023wall,lopez2023can,li2023chatgpt}.
Specifically, the most recent large language models (LLMs)~\citep{brown2020language}\footnote{\url{https://openai.com/blog/chatgpt}} have exhibited remarkable abilities in natural language understanding (NLU) and performing various tasks by following natural language instructions without training data.
Despite these successes, the highly technical nature of financial texts requires domain-specific LLMs to understand complex financial language and concepts effectively.
Such efforts include existing financial pre-trained language models (PLMs) such as finBERT~\citep{araci2019finbert}, FinBERT~\citep{yang2020finbert} and FLANG~\citep{shah2022flue}. However, those models are considered small since their parameter size is below one billion, limiting their generalization ability.
Recently, a proprietary financial LLM called BloombergGPT~\citep{wu2023bloomberggpt} with 50 billion parameters has been proposed by pre-training a Bloom-style LLM~\citep{scao2022bloom} on large-scale financial data. 

Despite these efforts, there remain several issues, as shown in Table \ref{tab:com}. Firstly, BloombergGPT and its training data are not openly released. Currently, there are no open-sourced financial LLMs, which can hinder development in the research community. Secondly, previous financial PLMs and the latest BloombergGPT are not fine-tuned for following natural language instructions (also known as instruction tuning), which is critical for improving the zero-shot ability on dealing with downstream financial tasks~\citep{wei2021finetuned,ouyang2022training}.
Thirdly, there are also no financial instruction data for supporting the instruction tuning of LLMs and evaluation benchmarks for comprehensively assessing and comparing the abilities of LLMs for financial tasks. 
We are thus motivated to consider the following research questions: 1) how can we develop efficient and openly available LLMs tailored for finance? 2) how can we build large-scale and high-quality financial instruction data? 3) how can we build the holistic financial evaluation benchmark for assessing financial LLMs?
\begin{table}[htb!]
    \centering
    \scriptsize
    \caption{The comparison of pre-trained language models and large language models for finance. "Instruct" means whether the model can follow instructions. "NLP" and "Fin" mean if the model is evaluated with financial NLP tasks and financial prediction tasks. }
    \label{tab:com}
    \begin{tabular}{llllllllll}
    \toprule
 \multirow{2}{*}{\textbf{Model}}&\multirow{2}{*}{\textbf{Backbone}}&\multirow{2}{*}{\textbf{Size}}&\multicolumn{2}{l}{\textbf{Open Source}}&\multirow{2}{*}{\textbf{Instruct}}&\multirow{2}{*}{\textbf{Language}}&\multicolumn{2}{l}{\textbf{Evaluation}}&\multirow{2}{*}{\textbf{Release Date}}\\
 & & & \textbf{Model} &\textbf{Data}&&&\textbf{NLP}&\textbf{Fin}&\\
finBERT~\citep{araci2019finbert}&BERT&110M&\Checkmark&\Checkmark&\XSolidBrush&English&\Checkmark&\XSolidBrush&08/27/19\\
FinBERT~\citep{yang2020finbert}&BERT&110M&\Checkmark&\XSolidBrush&\XSolidBrush&English&\Checkmark&\XSolidBrush&06/15/20\\
Mengzi-fin~\citep{zhang2021mengzi}&RoBERTa&103M&\Checkmark&\XSolidBrush&\XSolidBrush&Chinese&\Checkmark&\XSolidBrush&10/13/21\\
FLANG~\citep{shah2022flue}&ELECTRA&110M&\Checkmark&\Checkmark&\XSolidBrush&English&\Checkmark&\XSolidBrush&10/31/22\\
 BBT-FinT5~\citep{lu2023bbt}&T5&220M&\Checkmark&\Checkmark&\XSolidBrush&Chinese&\Checkmark&\XSolidBrush&02/18/23\\
 BloombergGPT~\citep{wu2023bloomberggpt}&BLOOM&50B&\XSolidBrush&\XSolidBrush&\XSolidBrush&English&\Checkmark&\XSolidBrush&03/30/23\\
FinMA&LLaMA&7/13B&\Checkmark&\Checkmark&\Checkmark&English&\Checkmark&\Checkmark&06/01/23\\
 \bottomrule
    \end{tabular}
\end{table}

\begin{figure}[t]
    \centering
    \includegraphics[width=0.9\textwidth]{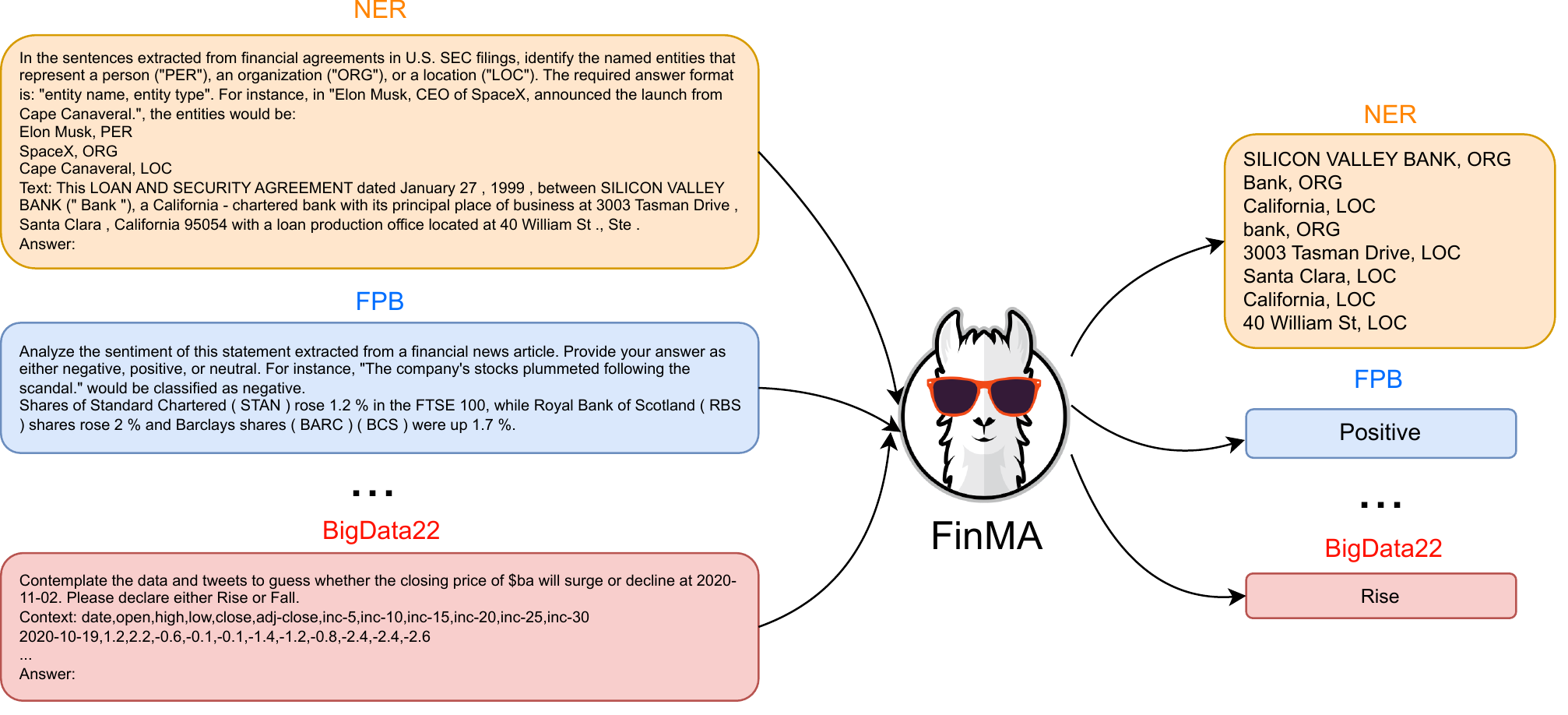}
    \caption{An overview of multi-task and multi-modal instruction tuning of FinMA for diverse financial tasks.}
    \label{fig:FinMA}
\end{figure}

\begin{CJK}{UTF8}{gkai}
To deal with these research questions, we propose PIXIU (貔貅)\footnote{PIXIU (貔貅)\url{https://en.wikipedia.org/wiki/Pixiu} is a mythical creature in Chinese folklore. It has the head of a dragon and the body of a lion and is believed to be an auspicious creature attracting money and good fortune.}, a comprehensive framework that includes the first financial LLM, FinMA, based on fine-tuning LLaMA~\citep{touvron2023llama} with multi-task and multi-modal instruction data. Fig~\ref{fig:FinMA} presents an overview of multi-task and multi-modal instruction tuning of FinMA for diverse
financial tasks. PIXIU also contains the first instruction data with 136K data samples to support the fine-tuning and a holistic evaluation benchmark with four financial NLP tasks and one financial prediction task. It has the following distinguishing features:
\begin{itemize}
    \item \textbf{Open resources}. We have openly released the financial LLM, instruction tuning data, and datasets included in the evaluation benchmark, and implementation, to encourage open research and transparency in the research field.
    \item \textbf{Multi-task}. PIXIU includes multi-task instruction tuning data covering a diverse set of financial tasks, including four financial NLP tasks and one financial prediction task.
    The multi-task instruction tuning has been proven to be critical for improving the model's generalization ability~\citep{sanh2022multitask,longpre2023flan} to new tasks.
     \item \textbf{Multi-modality}. Our instruction tuning data consists of multi-modality financial data such as tables in financial reports and historical stock prices as time-series data for the stock-movement prediction tasks beyond texts. Moreover, they encompass diverse types of financial texts, including reports, news articles, tweets, and regulatory filings. 
    \item \textbf{Diversity}. Compared with the evaluation tasks used in BloombergGPT and existing FLUE benchmark~\citep{shah2022flue}, which mainly cover financial NLP tasks, our evaluation benchmark includes stock movement prediction tasks.
    It requires the model to fully exploit both natural texts and time-series data to extract essential information for accurate prediction.
    Compared with financial NLP tasks, the financial prediction task is more aligned with real-world scenarios and more challenging.
\end{itemize}

To build the multi-task and multi-modal instruction data, we collect open-released training data from diverse tasks, including financial sentiment analysis, news headline classification, named entity recognition, question answering, and stock movement prediction, and propose the diverse task-specific instructions written by domain experts for each task. We create a large-scale instruction tuning data \textbf{FIT} by assembling the task-specific instructions with data samples from each task. We thus propose the domain-specific LLM \textbf{FinMA} by conducting the multi-task instruction tuning on LLaMA with the building dataset. 
To evaluate our model and other LLMs holistically, we build the \textbf{F}inancial \textbf{L}anguage Understanding \textbf{A}nd P\textbf{R}ediction \textbf{E}valuation Benchmark (FLARE) covering 4 financial NLP tasks with 6 datasets, and 1 financial prediction task with 3 datasets. 

Based on FLARE, we evaluate the performance of our model, BloombergGPT, and advanced LLMs in the general domain, such as ChatGPT\footnote{\url{https://openai.com/blog/chatgpt}} and GPT-4~\citep{openai2023gpt4}. Experimental results show that: 1) FinMA significantly outperforms LLMs, including BloombergGPT, ChatGPT, and GPT-4 on most tasks in FLARE, including financial sentiment analysis, news headline classification, NER, and stock movement prediction. This demonstrates the importance of tailoring the LLMs specifically for the financial domain. 2) Despite promising results on most tasks, FinMA underperforms BloombergGPT, ChatGPT, and GPT-4 on the question answering, which assesses the quantitative reasoning ability of LLMs. Our analysis finds that this is caused by the limitation of LLaMA on quantitative reasoning and mathematics. 3) Compared with NLP tasks, all LLMs, including FinMA, ChatGPT and GPT-4, still present limited performance on stock movement prediction, indicating room for further improvement. 4) FinMA fine-tuned with both NLP and financial prediction tasks, presents the best performance on one of the stock prediction datasets, indicating the potential of task-specific instruction tuning of LLMs on financial prediction tasks.

Our contributions can be summarized as follows: 1) We introduce FIT, the first multi-task and multi-modal instruction tuning data in the financial domain, covering 5 tasks and 9 datasets with 136,609 (136K) data samples. 
2) We introduce FLARE, the first evaluation benchmark with both financial natural language understanding and prediction tasks.
3) We introduce FinMA, the first openly released and instruction-following financial large language model, which achieves SOTA on 3 financial NLP tasks and 1 financial prediction task.
4) We compare FinMA and existing LLMs on FLARE. The results demonstrate the superiority of FinMA, the key limitations of LLMs for finance, and future directions to advance LLMs for finance.

\end{CJK}

\section{Related Work}
\label{sec:related}
\textbf{Financial Language Models}
Many PLMs for the financial domain have been proposed by continual pre-training PLMs with large-scale financial texts.
\cite{araci2019finbert} proposed the first financial PLM called finBERT that pre-trained BERT~\citep{kenton2019bert} with open released financial corpus such as TRC2-financial\footnote{\url{https://trec.nist.gov/data/reuters/reuters.html}} and Financial Phrase Bank~\citep{malo2014good}. finBERT outperforms neural network methods such as LSTM in financial sentiment classification tasks.
\cite{yang2020finbert} further proposed FinBERT by pre-training BERT with a 4.9 billion tokens financial communication corpus, which outperforms BERT on three financial sentiment classification datasets.
\cite{shah2022flue} proposed FLANG, a financial PLM with BERT and ELECTRA~\citep{clark2020electra} as the backbone.
Besides English, financial PLMs in other languages, such as Chinese, were also proposed, such as Mengzi-fin~\citep{zhang2021mengzi} and BBT-FinT5~\citep{lu2023bbt}. 
Latest, \cite{wu2023bloomberggpt} proposed BloombergGPT, the first financial large language model with 50 billion parameters, that is pre-trained with mixed datasets from the general and financial domain.
However, neither the model nor pre-trained domain datasets are not released. The model is also not instruction-following like other LLMs such as ChatGPT and GPT-4.

\textbf{Financial Evaluation Benchmark}
\cite{shah2022flue} proposed the first heterogeneous evaluation benchmark FLUE with 5 financial NLP tasks, including financial sentiment analysis~\citep{malo2014good}, news headline classification~\citep{sinha2021impact}, named entity recognition~\citep{alvarado2015domain}, structure boundary detection \footnote{\url{https://sites.google.com/nlg.csie.ntu.edu.tw/finweb2021/shared-task-finsbd-3}} and question answering~\citep{maia201818}.
\cite{lu2023bbt} proposed the first Chinese financial evaluation benchmark BBT-CFLEB \footnote{\url{https://bbt.ssymmetry.com/evaluation.html}} with financial news classification, summarization, relation extraction, question answering, and negative news determination task, as well as sentiment classification task of financial social media texts.
However, these benchmarks only consider financial NLP tasks and don't include financial prediction tasks, such as stock movement prediction~\citep{soun2022accurate} or pair trading~\citep{Han2023SelectAT} that are critical for evaluating the model's performance applied to real-world scenarios.

\textbf{Open Sourced Large Language Models}
Recent studies have made efforts on democratic AI, where the representative work is LLaMA~\citep{touvron2023llama} from Meta AI, an open-source LLM with parameters ranging from 7B and 13B to 65B. LLaMA-13B has comparable and even better performance than GPT-3~\citep{brown2020language} with 175B parameters on common sense reasoning tasks.
Following efforts have been proposed to improve LLaMA for instruction following like ChatGPT, by instruction tuning.
Such as \cite{alpaca} proposed Alpaca by fine-tuning LLaMA-7B with 52K instruction-following samples generated with the self-instruct method~\citep{wang2022self}. 
\cite{vicuna2023} proposed Vicuna-13B by fine-tuning LLaMA-13B with 70K conversation data from ShareGPT \footnote{\url{https://sharegpt.com}}. It can generate better answers to user's questions compared with Alpaca.
However, there are no open-sourced LLMs and instruction-tuning data focused on the financial domain.

\section{FIT: Financial Instruction Tuning Dataset}
In this section, we introduce our financial instruction tuning dataset FIT, including the background of raw data, tasks in FIT, and the construction process based on raw data. Different from existing financial datasets, FIT is the first instruction-tuning dataset for finance LLMs and includes stock movement prediction except for financial NLP tasks, which is fundamental for real-world financial applications.
\subsection{Raw Data}
\label{sec:raw-data}
Derived from real-world finance scenarios, we build our financial instruction tuning dataset FIT based on the open-sourced data of various financial NLP and prediction tasks.
Compared with the self-instruct method~\citep{wang2022self} commonly used by existing LLMs such as Alpaca, we choose to build instruction tuning datasets from open-sourced datasets due to the following reasons: 1) the open-sourced datasets are usually annotated by domain experts, showing high quality, 2) it has very low cost and has no limitation on commercial use unlike datasets constructed from ChatGPT or GPT-4, 3) these open-sourced datasets cover a variety of text types such as news, reports and tweets, as well as multi-modalities including time series data, tables, and texts. The details\footnote{For further details of the data split and pre-processing, please refer to Appendix} of the raw data and instruction data are shown in Table \ref{tab:raw-data}.
\begin{table}[htb!]
    \centering
    \scriptsize
    \caption{The details of the raw data and instruction data.}
    \label{tab:raw-data}
    \begin{tabular}{lllllll}
    \toprule
 \textbf{Data}&\textbf{Task}&\textbf{Raw}&\textbf{Instruction}&\textbf{Data Types}&\textbf{Modalities}&\textbf{License}\\
    \midrule
FPB&sentiment analysis&4,845&48,450&news&text&CC BY-SA 3.0\\
FiQA-SA&sentiment analysis&1,173&11,730&news headlines,tweets&text&Public\\
Headline&news headline classification&11,412&11,412&news headlines&text&CC BY-SA 3.0\\
NER&named entity recognition&1,366&13,660&financial agreements&text&CC BY-SA 3.0\\
FinQA&question answering&8,281&8,281&earnings reports&text,table&MIT License\\
ConvFinQA&question answering&3,892&3,892&earnings reports&text,table&MIT License\\
BigData22&stock movement prediction&7,164&7,164&tweets,historical prices&text,time series&Public\\
ACL18&stock movement prediction&27,053&27,053&tweets,historical prices&text,time series&MIT License\\
CIKM18&stock movement prediction&4,967&4,967&tweets,historical prices&text,time series&Public\\
 \bottomrule
    \end{tabular}
\end{table}

\textbf{Financial Sentiment Analysis.}
Financial sentiment analysis task has long been a critical task in the financial domain~\citep{araci2019finbert,yang2020finbert}, aiming to analyze the sentiment information of the input financial texts. Following existing benchmark FLUE~\citep{shah2022flue}, we use two datasets: the Financial Phrase Bank (FPB) dataset~\citep{malo2014good} and FiQA-SA~\citep{maia201818}. FPB includes English sentences from financial news and their sentiment label of positive, negative, or neutral annotated by domain experts. FiQA-SA is another widely-adopted dataset, which aims to predict the sentiment of English financial news and microblog posts on a scale of [-1,1], where 1 means the most positive.

\textbf{News Headline Classification.} 
The news headline classification task aims to analyze other information, such as price movement in financial texts. We use the Gold news headline dataset~\citep{sinha2021impact} consisting of news headlines from 2000 to 2019 about "gold" and their corresponding 9 tags: “price or not”, “price up”, “price down”, “price stable”, “past price”, “future price”, “past general”, “future general”, “asset comparison”. The task is to conduct the binary classification for each tag of each data sample.

\textbf{Named Entity Recognition.}
Named Entity Recognition (NER) task is to detect critical financial entities such as persons, organizations, and locations, which can be used to build financial knowledge graphs. We use the FIN dataset~\citep{alvarado2015domain} including sentences from public financial agreements through U.S. Security and Exchange Commission (SEC) filings and manually annotated entity types from LOCATION (LOC), ORGANISATION (ORG) and PERSON (PER).

\textbf{Question Answering.}
Question answering is the task of automatically answer a financial question based on the provided information. We use two datasets: FinQA~\citep{chen2021finqa} and ConvFinQA~\citep{chen2022convfinqa}. FinQA consists of question-answering pairs annotated by experts and their corresponding earnings reports (including unstructured documents and tables) from S\&P 500 companies. ConvFinQA is an expansion on FinQA that has conversations with the multi-turn question and answering over earnings reports.

\textbf{Stock Movement Prediction.}
As one of the fundamental financial tasks, stock movement prediction has great potential value in real applications such as investment strategies.
Following previous work~\citep{soun2022accurate}, we frame the task as a binary classification problem, which is to predict the binary stock price movement given historical stock prices and tweets.
If price movement is higher than 0.55\%, it will be assigned to positive samples (1), or negative samples (-1) if it is lower than -0.5\%.
We adopt three commonly-used datasets: BigData22~\citep{soun2022accurate}, ACL18~\citep{xu2018stock}, and CIKM18~\citep{wu2018hybrid}.

\subsection{Instruction construction}
\begin{table}[htb!]
    \centering
    \scriptsize
    \caption{The example prompt for each dataset. FiQA-SA has two types of text, including news headlines and tweets. We will fill the detailed text type into \textcolor{blue}{\{category\}} for each data sample. For stock movement prediction data such as BigData22, we will fill \textcolor{blue}{\{tid\}} and \textcolor{blue}{\{point\}} with the detailed stock name and time from each data sample.}
    \label{tab:prompt}
    \begin{tabular}{ll}
    \toprule
    \textbf{Data} & \textbf{Prompt} \\
    \midrule
    FPB & \makecell[l]{``Analyze the sentiment of this statement extracted from a financial news article.\\Provide your answer as either negative, positive or neutral.\\For instance, 'The company's stocks plummeted following the scandal.' would be classified as negative."} \\
    \midrule
    FiQA-SA & \makecell[l]{``What is the sentiment of the following financial \textcolor{blue}{\{category\}}:\\Positive, Negative, or Neutral?"} \\
    \midrule
    Headline & \makecell[l]{``Consider whether the headline mentions the price of gold.\\Is there a Price or Not in the gold commodity market indicated in the news headline?\\Please answer Yes or No."} \\
    \midrule
    NER & \makecell[l]{``In the sentences extracted from financial agreements in U.S. SEC filings,\\identify the named entities that represent a person ('PER'), an organization ('ORG'),\\or a location ('LOC'). The required answer format is: 'entity name, entity type'.\\For instance, in 'Elon Musk, CEO of SpaceX, announced the launch from Cape Canaveral.',\\the entities would be: 'Elon Musk, PER; SpaceX, ORG; Cape Canaveral, LOC'"} \\
    \midrule
    FinQA & \makecell[l]{``Given the financial data and expert analysis, please answer this question:"} \\
    \midrule
    ConvFinQA & \makecell[l]{``In the context of this series of interconnected finance-related queries and the additional information \\provided by the pretext, table data, and post text from a company's financial filings,\\please provide a response to the final question. This may require extracting information\\ from the context and performing mathematical calculations. Please take into account the information provided in\\ the preceding questions and their answers when formulating your response:"} \\
    \midrule
    BigData22 & \makecell[l]{``Analyze the information and social media posts to determine if the closing price of \textcolor{blue}{\{tid\}}\\will ascend or descend at \textcolor{blue}{\{point\}}. Please respond with either Rise or Fall."} \\
    \bottomrule
    \end{tabular}
\end{table}

Base on the raw datasets,
we further construct our financial instruction datasets, whose statistics are presented in Table \ref{tab:raw-data}.
We ask domain experts to write 10 diverse instructions for all datasets except the ConvFinQA, where we only use one instruction.
Since ConvFinQA is a multi-turn conversational question-answering dataset, which has diverse questions as instructions in nature. For BigData22, ACL18, CIKM18, we use the same instruction set, since they have the same data types of input data and task formulation.
We show the instruction examples in Table \ref{tab:prompt}.
Based on these prompts, we convert raw datasets from these tasks into instruction-tuning samples, by gathering human-designed instructions, and input texts along with responses of each dataset. For FPB, FiQA-SA, Headline, NER, BigData22, ACL18, and CIKM18 datasets, we build instruction tuning samples with the following template:

\colorbox[gray]{0.95}{Instruction: [task prompt]\quad
Text: [input text]\quad
Response: [output]
}

[task prompt] is the prompt designed for each data, [input text] is the input financial data from each data, e.g. the historical prices and tweets for stock movement prediction datasets, [output] is the corresponding output for input text, e.g. sentiment label of input text from ["Positive", "Negative", "Neutral"] in FiQA-SA dataset.
For FPB, FiQA-SA, and NER, due to the limited data size, we employ all 10 instructions for each sample, while we randomly sample one instruction for each sample in Headline, BigData22, ACL18, and CIKM18 datasets.

For FinQA and ConvFinQA, we use the following template:

\colorbox[gray]{0.95}{
\parbox{\textwidth}{
Instruction: [task prompt]\quad
Context: [input context]\quad
Question: [input question]\quad
Response: [answer]
}}
[input context] is the input contextual information for each data sample. For example, the input context can be filled with the text and table from the filling files for FinQA.
ConvFinQA has multi-turn conversations with questions and answering. We thus use the following template:
We transform each turn of the conversation for each data sample into one instruction via the template, which will append previous questions and answer in the [input context].

\section{FinMA: Financial Large Language Model}
\label{sec:finma}
We further build FinMA by fine-tuning LLaMA~\citep{touvron2023llama} with FIT.
We train three models: FinMA-7B and FinMA-30B by fine-tuning LLaMA 7B and 30B checkpoint with instruction tuning data covering NLP tasks, and FinMA-7B-full by fine-tuning LLaMA 7B with full instruction tuning data . 
We fine-tune LLaMA-7B  with 15 epochs and LLaAM-7B-full with 3 epochs based on AdamW optimizer~\citep{loshchilov2017decoupled}. The batch size is set to 32, the initial learning rate is 8e-6, and the weight decay is 1e-5. We also set warmup steps to 5\% of all training steps. The maximum length of input texts is 2048. The FinMA-7B is fine-tuned on 8 A100 40GB GPUs.
As for the FinMA-30B model, we fine-tune LLaMA-30B with 20 epochs, which is also based on the AdamW optimizer.
The batch size is set to 24, the initial learning rate is 8e-6, the weight decay is 1e-5, and warmup steps to 5\% of all training steps. The maximum length of input texts is 2048. Different from FinMA-7B, it can only be distributed fine-tuned on 128 A100 40GB GPUs.

\section{FLARE: Financial Evaluation Benchmark}
\label{sec:benchmark}
Based on FIT, we design our financial natural language understanding and prediction evaluation benchmark (FLARE). We randomly select validation sets from FIT to select the best model checkpoint, and test sets for evaluation. 
Compared with the existing benchmark FLUE~\citep{sanh2022multitask}, FLARE covers financial prediction tasks in addition to NLP tasks\footnote{Following BloombergGPT, we don't include the structure boundary detection task included in FLUE because they are hard to be converted into the instruction following task.}. We believe it is vital to include financial prediction tasks such as stock movement prediction, to comprehensively evaluate the performance of LLMs on the practical applications of the financial domain. We show the data statistics of validation, and test set for each dataset in Table \ref{tab:eval}. 
\begin{table}[htb!]
    \centering
    \scriptsize
     \caption{The details of our evaluation datasets. To compare the performance with BloombergGPT whose test data is not openly-released, we keep the same numbers and data distributions of our test datasets with that of BloombergGPT.}
    \label{tab:eval}
    \begin{tabular}{lllll}
    \toprule
 \textbf{Data}&\textbf{Task}&\textbf{Valid}&\textbf{Test}&\textbf{Evaluation}\\
    \midrule
FPB~\citep{malo2014good}&sentiment analysis&7,740&9,700&F1, Accuracy\\
FiQA-SA~\citep{maia201818}&sentiment analysis&1,880&2,350&F1, Accuracy\\
Headline~\citep{sinha2021impact}&news headline classification&1,0259&2,0547&Avg F1\\
NER~\citep{alvarado2015domain}&named entity recognition&1,029&980&Entity F1\\
FinQA~\citep{chen2021finqa}&question answering&882&1,147&EM Accuracy\\
ConvFinQA~\citep{chen2022convfinqa}&question answering&1,489&2,161&EM Accuracy\\
BigData22~\citep{soun2022accurate}&stock movement prediction&797&1,471&Accurady, MCC\\
ACL18~\citep{xu2018stock}&stock movement prediction&2,554&3,719&Accurady, MCC\\
CIKM18~\citep{wu2018hybrid}&stock movement prediction&430&1,142&Accurady, MCC\\
 \bottomrule
    \end{tabular}
\end{table}
Following previous methods~\citep{li2023chatgpt,shah2022flue}, we evaluate the performance of the sentiment classification task on FPB and FiQA-SA datasets, with the accuracy (ACC) and weighted F1 Score (F1). 
The performance of the news headline classification task is evaluated with the weighted averages of F1 score over all nine categories (Avg F1). 
For the performance of NER task, we evaluate with the entity-level F1 score (Entity F1). The performance on the question-answering task is evaluated with the exact match accuracy (EM Acc).
As for the financial prediction task, following previous methods~\citep{xu2018stock,xie2023wall}, we evaluate the performance with the accuracy (ACC) and the Matthews correlation coefficient (MCC).

\section{Experiments on FLARE}
\label{sec:experiments}
The proposed FIT and FLARE allow to train, select the model, and evaluate the performance of LLMs on financial understanding and predictions.
In this section, we investigate how powerful the FIT-fine-tuned FinMA and other LLMs are on FLARE.
We compare FinMA with following LLMs: 1) BloombergGPT~\citep{wu2023bloomberggpt}. The only large language model with 50B parameters pre-trained with the financial texts.
2) GPT-4~\citep{openai2023gpt4}. A powerful instruction following large language model with around 1T parameters proposed by OpenAI. 
3) ChatGPT. A instruction following large language model with 175B parameters from OpenAI.
4) BLOOM~\citep{scao2022bloom}. An open-access multilingual large language model with 176B parameters.
5) GPT-NeoX~\citep{gpt-neox-20b}. An open-sourced large language model with 20B parameters.
6) OPT-66B~\citep{zhang2022opt}. An open-sourced language model with parameters from 125M to 175B. We use the OPT with 66B parameters.
7) Vicuna-13B~\citep{zhang2022opt}. An instruction following large language model by fine-tuning LLaMA-13B.

Following previous methods~\citep{wu2023bloomberggpt,li2023chatgpt}, we report the 20-shot performance of BloombergGPT and the 5-shot performance of other baseline methods on the FIN dataset. We report the 5-shot performance of BloombergGPT on FPB and FiQA-SA. We report the 5-shot performance of all baselines on the News dataset. For the remaining results, we report the zero-shot performance.
The results of some baselines are based on human evaluations, since LLMs without fine-tuning will fail to generate answers pre-defined in the given instruction.
All results of FinMA are conducted on zero-shot and can be automatically evaluated.

\begin{table}[htb!]
    \centering
    \scriptsize
\caption{The zero-shot and few-shot performance of different LLMs on the FLARE benchmark. Some results are referenced from~\citep{wu2023bloomberggpt,li2023chatgpt,xie2023wall}. BloombergGPT doesn't release their test datasets. Test datasets were built to have the same data distribution with that of BloombergGPT and the performance of FinMA was directly compared with BloombergGPT following the previous method~\citep{li2023chatgpt}.}
    \label{tab:per}
\begin{tabular}{l|l|ccccccccc}
    \toprule
    \textbf{Dataset} &\textbf{Metrics}&\makecell{\textbf{GPT}\\\textbf{NeoX}}&\makecell{\textbf{OPT}\\\textbf{66B}}&\textbf{BLOOM}&\makecell{\textbf{Chat}\\\textbf{GPT}}&\makecell{\textbf{GPT}\\\textbf{4}}&\makecell{\textbf{Bloomberg}\\\textbf{GPT}}&\makecell{\textbf{FinMA}\\\textbf{7B}}&\makecell{\textbf{FinMA}\\\textbf{30B}}&\makecell{\textbf{FinMA}\\\textbf{7B-full}}\\
    \midrule
    \multirow{2}{*}{FPB}&Acc&-&-&-&0.78&0.76&-&0.86&\textbf{0.87}&0.87\\\cline{2-11}
    &F1&0.45&0.49&0.50&0.78&0.78&0.51&0.86&\textbf{0.88}&0.87\\\cline{1-11}
    FiQA-SA&F1&0.51&0.52&0.53&-&-&0.75&0.84&\textbf{0.87}&0.79\\\hline
    Headline &AvgF1&0.73&0.79&0.77&0.77&0.86&0.82&\textbf{0.98}&0.97&0.97\\\hline
    NER&EntityF1&0.61&0.57&0.56&0.77&\textbf{0.83}&0.61&0.75&0.62&0.69\\\hline
    FinQA&EmAcc&-&-&-&0.58&\textbf{0.63}&-&0.06&0.11&0.04\\\hline
    ConvFinQA&EmAcc&0.28&0.30&0.36&0.60&\textbf{0.76}&0.43&0.25&0.40&0.20\\\hline
    \multirow{2}{*}{BigData22}&Acc&-&-&-&0.53&\textbf{0.54}&-&0.48&0.47&0.49\\\cline{2-11}
    &MCC&-&-&-&-0.025&0.03&-&0.04&\textbf{0.04}&0.01\\\hline
    \multirow{2}{*}{ACL18}&Acc&-&-&-&0.50&0.52&-&0.50&0.49&\textbf{0.56}\\\cline{2-11}
    &MCC&-&-&-&0.005&0.02&-&0.00&0.00&\textbf{0.10}\\\hline
    \multirow{2}{*}{CIKM18}&Acc&-&-&-&0.55&\textbf{0.57}&-&0.56&0.43&0.53\\\cline{2-11}
    &MCC&-&-&-&0.01&\textbf{0.02}&-&-0.02&-0.05&-0.03\\
    \bottomrule
\end{tabular}
\end{table}

\subsection{Results}
\label{sec:results}
\textbf{Overall Performance.}
For financial NLP tasks, as shown in Table \ref{tab:per}, our fine-tuned model FinMA significantly outperform other LLMs on FPB, FiQA-SA and Headline datasets, showing the importance of domain specific instruction tuning on improving the performance of LLMs in the specific domain.
For example, FinMA-30B outperforms GPT-4 by 10\% F1 score, and BloombergGPT by 37\% F1 score on the FPB dataset.
On the NER dataset, FinMA-7B also outperforms BloombergGPT and other LLMs, and achieve competitive results compared with ChatGPT and GPT-4.
Yet for FinQA and ConvFinQA which requires complex numeric reasoning, there is a large gap between the performance of GPT and FinMA.
As reported in existing studies~\citep{touvron2023llama,lewkowycz2022solving}, LLaMA includes no mathematical datasets for pre-training, resulted in poor performance on the 
mathematical benchmark datasets such as GSM8K~\citep{cobbe2021training}.
This finding indicate the importance of numeric reasoning for financial question answering, which could be the potential direction for advancing LLMs in the finance area.
For financial prediction tasks, all LLMs including FinMA, ChatGPT and GPT-4 struggle in stock movement prediction. After fine-tuned with both NLP and financial prediction tasks, FinMA-7B-full can achieve a significantly better performance on ACL18 dataset compared with ChatGPT and GPT-4. However, it still presents almost zero MCC on the other two datasets like ChatGPT and GPT-4. This indicates the complexity and challenging of the financial prediction tasks in FLARE. Compared with existing financial benchmarks that focusing on NLP tasks, FLAPE provides exciting opportunities for the improvement of LLMs on the fundamental of financial academic studies and applications.

\textbf{Further Analysis.} We further analyze the influences of model size, and instruction tuning data on the performance of LLMs on different tasks. 
FinMA-30B has no significantly better performance than FinMA-7B on most NLP tasks and the stock movement prediction task.  
Apparently, the quality of the instructions rather than the model size is critical for the performance of these tasks.
For the complex question-answering tasks such as ConvFinQA, as shown in Table \ref{tab:conv}, the larger LLaMA model generally has better performance.
Particularly, Vicuna-7B based on LLaMA-7B has the worst performance, which are also consistent with previous findings~\citep{cobbe2021training} that LLaMA with larger parameters presents better performance on mathematical benchmark datasets.
FinMA-7B and FinMA-30B are not fine-tuned with the financial prediction dataset, they show limited performance on stock movement prediction similar to ChatGPT and GPT-4. Although they can be generalized to the unseen financial prediction task, there is a large room to be improved along with ChatGPT and GPT-4.
In contrast, FinMA-7B-full fine-tuned with both NLP and prediction datasets, has shown significantly better performance on the ACL18 dataset, and comparable performance on NLP tasks with FinMA-7B and GPT-4.
This indicates the potential of LLMs to be further adapted and applied directly on financial prediction tasks via pre-training and fine-tuning on domain datasets.
\begin{table}[htb!]
    \centering
    \scriptsize
\caption{The performance of LLMs over ConvFinQA.}
    \label{tab:conv}
\begin{tabular}{l|l|ccccccc}
    \toprule
    \textbf{Metrics}&\textbf{GPT-4}&\textbf{BloombergGPT}&\textbf{FinMA-7B}&\textbf{FinMA-30B}&\textbf{FinMA-7B-full}&\textbf{Vicuna-7B}\\
    \midrule
     EmAcc&\textbf{0.76}&0.43&0.25&0.40&0.20&0.10\\
    \bottomrule
\end{tabular}
\end{table}

\section{Limitations}
\label{sec:limitation}
Despite the positive contributions of this study, we recognize the following limitations:

\begin{enumerate}
    \item \textbf{Model and Training Constraints}: We only present FinMA models up to 30B. Due to computational constraints, FinMA-30B has not been fine-tuned on the full dataset.

    \item \textbf{Complex Task Performance}: FinMA, due to the limitation of the backbone model LLaMA, struggles with tasks requiring quantitative reasoning, such as financial question answering, and the difficult financial prediction task.

    \item \textbf{Resource Constraints and Generalizability}: The development of FinMA, FIT, and FLARE is influenced by available resources and handcrafted instructions, potentially affecting model diversity and generalizability. The maximum input size of FinMA is also limited by the maximum input texts that can be handled by the backbone model LLaMA.

    \item  \textbf{Potential Negative Impacts}: While our study primarily focuses on the positive aspects and advancements of financial language understanding models, it is important to acknowledge the potential negative impacts associated with their use, such as the spread of financial misinformation or unethical market influence.
\end{enumerate}

\section{Conclusion}
 In this work, we presented PIXIU, encompassing the first open-sourced financial large language model FinMA, the instruction tuning dataset FIT, and the evaluation benchmark FLARE. Through extensive evaluation, we demonstrated the effectiveness of FinMA across various financial tasks, showing the potential of domain-specific instruction tuning of large language models in the financial domain. However, challenges such as improving performance on complex tasks and addressing resource constraints remain. Our open-source contribution aims to facilitate further research and innovation in financial language understanding, prediction, and LLMs, toward more useful and safe LLMs in the field of finance.

\section*{Acknowledgement}
This paper would not have been possible without the invaluable contribution of the BELLE code \citep{BELLE, belle2023exploring}.
\bibliographystyle{ACM-Reference-Format}
\bibliography{pixiu}


\begin{thebibliography}{40}


\ifx \showCODEN    \undefined \def \showCODEN     #1{\unskip}     \fi
\ifx \showDOI      \undefined \def \showDOI       #1{#1}\fi
\ifx \showISBNx    \undefined \def \showISBNx     #1{\unskip}     \fi
\ifx \showISBNxiii \undefined \def \showISBNxiii  #1{\unskip}     \fi
\ifx \showISSN     \undefined \def \showISSN      #1{\unskip}     \fi
\ifx \showLCCN     \undefined \def \showLCCN      #1{\unskip}     \fi
\ifx \shownote     \undefined \def \shownote      #1{#1}          \fi
\ifx \showarticletitle \undefined \def \showarticletitle #1{#1}   \fi
\ifx \showURL      \undefined \def \showURL       {\relax}        \fi
\providecommand\bibfield[2]{#2}
\providecommand\bibinfo[2]{#2}
\providecommand\natexlab[1]{#1}
\providecommand\showeprint[2][]{arXiv:#2}

\bibitem[Alvarado et~al\mbox{.}(2015)]%
        {alvarado2015domain}
\bibfield{author}{\bibinfo{person}{Julio Cesar~Salinas Alvarado},
  \bibinfo{person}{Karin Verspoor}, {and} \bibinfo{person}{Timothy Baldwin}.}
  \bibinfo{year}{2015}\natexlab{}.
\newblock \showarticletitle{Domain adaption of named entity recognition to
  support credit risk assessment}. In \bibinfo{booktitle}{\emph{Proceedings of
  the Australasian Language Technology Association Workshop 2015}}.
  \bibinfo{pages}{84--90}.
\newblock


\bibitem[Araci(2019)]%
        {araci2019finbert}
\bibfield{author}{\bibinfo{person}{Dogu Araci}.}
  \bibinfo{year}{2019}\natexlab{}.
\newblock \showarticletitle{Finbert: Financial sentiment analysis with
  pre-trained language models}.
\newblock \bibinfo{journal}{\emph{arXiv preprint arXiv:1908.10063}}
  (\bibinfo{year}{2019}).
\newblock


\bibitem[Black et~al\mbox{.}(2022)]%
        {gpt-neox-20b}
\bibfield{author}{\bibinfo{person}{Sid Black}, \bibinfo{person}{Stella
  Biderman}, \bibinfo{person}{Eric Hallahan}, \bibinfo{person}{Quentin
  Anthony}, \bibinfo{person}{Leo Gao}, \bibinfo{person}{Laurence Golding},
  \bibinfo{person}{Horace He}, \bibinfo{person}{Connor Leahy},
  \bibinfo{person}{Kyle McDonell}, \bibinfo{person}{Jason Phang},
  \bibinfo{person}{Michael Pieler}, \bibinfo{person}{USVSN~Sai Prashanth},
  \bibinfo{person}{Shivanshu Purohit}, \bibinfo{person}{Laria Reynolds},
  \bibinfo{person}{Jonathan Tow}, \bibinfo{person}{Ben Wang}, {and}
  \bibinfo{person}{Samuel Weinbach}.} \bibinfo{year}{2022}\natexlab{}.
\newblock \showarticletitle{{GPT-NeoX-20B}: An Open-Source Autoregressive
  Language Model}. In \bibinfo{booktitle}{\emph{Proceedings of the ACL Workshop
  on Challenges \& Perspectives in Creating Large Language Models}}.
\newblock
\urldef\tempurl%
\url{https://arxiv.org/abs/2204.06745}
\showURL{%
\tempurl}


\bibitem[Brown et~al\mbox{.}(2020)]%
        {brown2020language}
\bibfield{author}{\bibinfo{person}{Tom Brown}, \bibinfo{person}{Benjamin Mann},
  \bibinfo{person}{Nick Ryder}, \bibinfo{person}{Melanie Subbiah},
  \bibinfo{person}{Jared~D Kaplan}, \bibinfo{person}{Prafulla Dhariwal},
  \bibinfo{person}{Arvind Neelakantan}, \bibinfo{person}{Pranav Shyam},
  \bibinfo{person}{Girish Sastry}, \bibinfo{person}{Amanda Askell},
  {et~al\mbox{.}}} \bibinfo{year}{2020}\natexlab{}.
\newblock \showarticletitle{Language models are few-shot learners}.
\newblock \bibinfo{journal}{\emph{Advances in neural information processing
  systems}}  \bibinfo{volume}{33} (\bibinfo{year}{2020}),
  \bibinfo{pages}{1877--1901}.
\newblock


\bibitem[Chen et~al\mbox{.}(2021)]%
        {chen2021finqa}
\bibfield{author}{\bibinfo{person}{Zhiyu Chen}, \bibinfo{person}{Wenhu Chen},
  \bibinfo{person}{Charese Smiley}, \bibinfo{person}{Sameena Shah},
  \bibinfo{person}{Iana Borova}, \bibinfo{person}{Dylan Langdon},
  \bibinfo{person}{Reema Moussa}, \bibinfo{person}{Matt Beane},
  \bibinfo{person}{Ting-Hao Huang}, \bibinfo{person}{Bryan~R Routledge},
  {et~al\mbox{.}}} \bibinfo{year}{2021}\natexlab{}.
\newblock \showarticletitle{FinQA: A Dataset of Numerical Reasoning over
  Financial Data}. In \bibinfo{booktitle}{\emph{Proceedings of the 2021
  Conference on Empirical Methods in Natural Language Processing}}.
  \bibinfo{pages}{3697--3711}.
\newblock


\bibitem[Chen et~al\mbox{.}(2022)]%
        {chen2022convfinqa}
\bibfield{author}{\bibinfo{person}{Zhiyu Chen}, \bibinfo{person}{Shiyang Li},
  \bibinfo{person}{Charese Smiley}, \bibinfo{person}{Zhiqiang Ma},
  \bibinfo{person}{Sameena Shah}, {and} \bibinfo{person}{William~Yang Wang}.}
  \bibinfo{year}{2022}\natexlab{}.
\newblock \showarticletitle{Convfinqa: Exploring the chain of numerical
  reasoning in conversational finance question answering}.
\newblock \bibinfo{journal}{\emph{arXiv preprint arXiv:2210.03849}}
  (\bibinfo{year}{2022}).
\newblock


\bibitem[Chiang et~al\mbox{.}(2023)]%
        {vicuna2023}
\bibfield{author}{\bibinfo{person}{Wei-Lin Chiang}, \bibinfo{person}{Zhuohan
  Li}, \bibinfo{person}{Zi Lin}, \bibinfo{person}{Ying Sheng},
  \bibinfo{person}{Zhanghao Wu}, \bibinfo{person}{Hao Zhang},
  \bibinfo{person}{Lianmin Zheng}, \bibinfo{person}{Siyuan Zhuang},
  \bibinfo{person}{Yonghao Zhuang}, \bibinfo{person}{Joseph~E. Gonzalez},
  \bibinfo{person}{Ion Stoica}, {and} \bibinfo{person}{Eric~P. Xing}.}
  \bibinfo{year}{2023}\natexlab{}.
\newblock \bibinfo{title}{Vicuna: An Open-Source Chatbot Impressing GPT-4 with
  90\%* ChatGPT Quality}.
\newblock
\newblock
\urldef\tempurl%
\url{https://lmsys.org/blog/2023-03-30-vicuna/}
\showURL{%
\tempurl}


\bibitem[Clark et~al\mbox{.}(2020)]%
        {clark2020electra}
\bibfield{author}{\bibinfo{person}{Kevin Clark}, \bibinfo{person}{Minh-Thang
  Luong}, \bibinfo{person}{Quoc~V Le}, {and} \bibinfo{person}{Christopher~D
  Manning}.} \bibinfo{year}{2020}\natexlab{}.
\newblock \showarticletitle{Electra: Pre-training text encoders as
  discriminators rather than generators}.
\newblock \bibinfo{journal}{\emph{arXiv preprint arXiv:2003.10555}}
  (\bibinfo{year}{2020}).
\newblock


\bibitem[Cobbe et~al\mbox{.}(2021)]%
        {cobbe2021training}
\bibfield{author}{\bibinfo{person}{Karl Cobbe}, \bibinfo{person}{Vineet
  Kosaraju}, \bibinfo{person}{Mohammad Bavarian}, \bibinfo{person}{Mark Chen},
  \bibinfo{person}{Heewoo Jun}, \bibinfo{person}{Lukasz Kaiser},
  \bibinfo{person}{Matthias Plappert}, \bibinfo{person}{Jerry Tworek},
  \bibinfo{person}{Jacob Hilton}, \bibinfo{person}{Reiichiro Nakano},
  {et~al\mbox{.}}} \bibinfo{year}{2021}\natexlab{}.
\newblock \showarticletitle{Training verifiers to solve math word problems}.
\newblock \bibinfo{journal}{\emph{arXiv preprint arXiv:2110.14168}}
  (\bibinfo{year}{2021}).
\newblock


\bibitem[Han et~al\mbox{.}(2023a)]%
        {han2023select}
\bibfield{author}{\bibinfo{person}{Weiguang Han}, \bibinfo{person}{Boyi Zhang},
  \bibinfo{person}{Qianqian Xie}, \bibinfo{person}{Min Peng},
  \bibinfo{person}{Yanzhao Lai}, {and} \bibinfo{person}{Jimin Huang}.}
  \bibinfo{year}{2023}\natexlab{a}.
\newblock \showarticletitle{Select and Trade: Towards Unified Pair Trading with
  Hierarchical Reinforcement Learning}.
\newblock \bibinfo{journal}{\emph{arXiv preprint arXiv:2301.10724}}
  (\bibinfo{year}{2023}).
\newblock


\bibitem[Han et~al\mbox{.}(2023b)]%
        {Han2023SelectAT}
\bibfield{author}{\bibinfo{person}{Weiguang Han}, \bibinfo{person}{Boyi Zhang},
  \bibinfo{person}{Qianqian Xie}, \bibinfo{person}{Min Peng},
  \bibinfo{person}{Yanzhao Lai}, {and} \bibinfo{person}{Jimin Huang}.}
  \bibinfo{year}{2023}\natexlab{b}.
\newblock \showarticletitle{Select and Trade: Towards Unified Pair Trading with
  Hierarchical Reinforcement Learning}.
\newblock \bibinfo{journal}{\emph{ArXiv}}  \bibinfo{volume}{abs/2301.10724}
  (\bibinfo{year}{2023}).
\newblock


\bibitem[Ji et~al\mbox{.}(2023a)]%
        {BELLE}
\bibfield{author}{\bibinfo{person}{Yunjie Ji}, \bibinfo{person}{Yong Deng},
  \bibinfo{person}{Yan Gong}, \bibinfo{person}{Yiping Peng},
  \bibinfo{person}{Qiang Niu}, \bibinfo{person}{Baochang Ma}, {and}
  \bibinfo{person}{Xiangang Li}.} \bibinfo{year}{2023}\natexlab{a}.
\newblock \bibinfo{title}{BELLE: Be Everyone's Large Language model Engine}.
\newblock \bibinfo{howpublished}{\url{https://github.com/LianjiaTech/BELLE}}.
\newblock


\bibitem[Ji et~al\mbox{.}(2023b)]%
        {belle2023exploring}
\bibfield{author}{\bibinfo{person}{Yunjie Ji}, \bibinfo{person}{Yong Deng},
  \bibinfo{person}{Yan Gong}, \bibinfo{person}{Yiping Peng},
  \bibinfo{person}{Qiang Niu}, \bibinfo{person}{Lei Zhang},
  \bibinfo{person}{Baochang Ma}, {and} \bibinfo{person}{Xiangang Li}.}
  \bibinfo{year}{2023}\natexlab{b}.
\newblock \showarticletitle{Exploring the Impact of Instruction Data Scaling on
  Large Language Models: An Empirical Study on Real-World Use Cases}.
\newblock \bibinfo{journal}{\emph{arXiv preprint arXiv:2303.14742}}
  (\bibinfo{year}{2023}).
\newblock


\bibitem[Kenton and Toutanova(2019)]%
        {kenton2019bert}
\bibfield{author}{\bibinfo{person}{Jacob Devlin Ming-Wei~Chang Kenton} {and}
  \bibinfo{person}{Lee~Kristina Toutanova}.} \bibinfo{year}{2019}\natexlab{}.
\newblock \showarticletitle{BERT: Pre-training of Deep Bidirectional
  Transformers for Language Understanding}. In
  \bibinfo{booktitle}{\emph{Proceedings of NAACL-HLT}}.
  \bibinfo{pages}{4171--4186}.
\newblock


\bibitem[Lewkowycz et~al\mbox{.}(2022)]%
        {lewkowycz2022solving}
\bibfield{author}{\bibinfo{person}{Aitor Lewkowycz}, \bibinfo{person}{Anders
  Andreassen}, \bibinfo{person}{David Dohan}, \bibinfo{person}{Ethan Dyer},
  \bibinfo{person}{Henryk Michalewski}, \bibinfo{person}{Vinay Ramasesh},
  \bibinfo{person}{Ambrose Slone}, \bibinfo{person}{Cem Anil},
  \bibinfo{person}{Imanol Schlag}, \bibinfo{person}{Theo Gutman-Solo},
  {et~al\mbox{.}}} \bibinfo{year}{2022}\natexlab{}.
\newblock \showarticletitle{Solving quantitative reasoning problems with
  language models}.
\newblock \bibinfo{journal}{\emph{arXiv preprint arXiv:2206.14858}}
  (\bibinfo{year}{2022}).
\newblock


\bibitem[Li et~al\mbox{.}(2023)]%
        {li2023chatgpt}
\bibfield{author}{\bibinfo{person}{Xianzhi Li}, \bibinfo{person}{Xiaodan Zhu},
  \bibinfo{person}{Zhiqiang Ma}, \bibinfo{person}{Xiaomo Liu}, {and}
  \bibinfo{person}{Sameena Shah}.} \bibinfo{year}{2023}\natexlab{}.
\newblock \showarticletitle{Are ChatGPT and GPT-4 General-Purpose Solvers for
  Financial Text Analytics? An Examination on Several Typical Tasks}.
\newblock \bibinfo{journal}{\emph{arXiv preprint arXiv:2305.05862}}
  (\bibinfo{year}{2023}).
\newblock


\bibitem[Longpre et~al\mbox{.}(2023)]%
        {longpre2023flan}
\bibfield{author}{\bibinfo{person}{Shayne Longpre}, \bibinfo{person}{Le Hou},
  \bibinfo{person}{Tu Vu}, \bibinfo{person}{Albert Webson},
  \bibinfo{person}{Hyung~Won Chung}, \bibinfo{person}{Yi Tay},
  \bibinfo{person}{Denny Zhou}, \bibinfo{person}{Quoc~V Le},
  \bibinfo{person}{Barret Zoph}, \bibinfo{person}{Jason Wei}, {et~al\mbox{.}}}
  \bibinfo{year}{2023}\natexlab{}.
\newblock \showarticletitle{The flan collection: Designing data and methods for
  effective instruction tuning}.
\newblock \bibinfo{journal}{\emph{arXiv preprint arXiv:2301.13688}}
  (\bibinfo{year}{2023}).
\newblock


\bibitem[Lopez-Lira and Tang(2023)]%
        {lopez2023can}
\bibfield{author}{\bibinfo{person}{Alejandro Lopez-Lira} {and}
  \bibinfo{person}{Yuehua Tang}.} \bibinfo{year}{2023}\natexlab{}.
\newblock \showarticletitle{Can ChatGPT Forecast Stock Price Movements? Return
  Predictability and Large Language Models}.
\newblock \bibinfo{journal}{\emph{arXiv preprint arXiv:2304.07619}}
  (\bibinfo{year}{2023}).
\newblock


\bibitem[Loshchilov and Hutter(2017)]%
        {loshchilov2017decoupled}
\bibfield{author}{\bibinfo{person}{Ilya Loshchilov} {and}
  \bibinfo{person}{Frank Hutter}.} \bibinfo{year}{2017}\natexlab{}.
\newblock \showarticletitle{Decoupled weight decay regularization}.
\newblock \bibinfo{journal}{\emph{arXiv preprint arXiv:1711.05101}}
  (\bibinfo{year}{2017}).
\newblock


\bibitem[Lu et~al\mbox{.}(2023)]%
        {lu2023bbt}
\bibfield{author}{\bibinfo{person}{Dakuan Lu}, \bibinfo{person}{Jiaqing Liang},
  \bibinfo{person}{Yipei Xu}, \bibinfo{person}{Qianyu He},
  \bibinfo{person}{Yipeng Geng}, \bibinfo{person}{Mengkun Han},
  \bibinfo{person}{Yingsi Xin}, \bibinfo{person}{Hengkui Wu}, {and}
  \bibinfo{person}{Yanghua Xiao}.} \bibinfo{year}{2023}\natexlab{}.
\newblock \showarticletitle{BBT-Fin: Comprehensive Construction of Chinese
  Financial Domain Pre-trained Language Model, Corpus and Benchmark}.
\newblock \bibinfo{journal}{\emph{arXiv preprint arXiv:2302.09432}}
  (\bibinfo{year}{2023}).
\newblock


\bibitem[Maia et~al\mbox{.}(2018)]%
        {maia201818}
\bibfield{author}{\bibinfo{person}{Macedo Maia}, \bibinfo{person}{Siegfried
  Handschuh}, \bibinfo{person}{Andr{\'e} Freitas}, \bibinfo{person}{Brian
  Davis}, \bibinfo{person}{Ross McDermott}, \bibinfo{person}{Manel Zarrouk},
  {and} \bibinfo{person}{Alexandra Balahur}.} \bibinfo{year}{2018}\natexlab{}.
\newblock \showarticletitle{Www'18 open challenge: financial opinion mining and
  question answering}. In \bibinfo{booktitle}{\emph{Companion proceedings of
  the the web conference 2018}}. \bibinfo{pages}{1941--1942}.
\newblock


\bibitem[Malo et~al\mbox{.}(2014)]%
        {malo2014good}
\bibfield{author}{\bibinfo{person}{Pekka Malo}, \bibinfo{person}{Ankur Sinha},
  \bibinfo{person}{Pekka Korhonen}, \bibinfo{person}{Jyrki Wallenius}, {and}
  \bibinfo{person}{Pyry Takala}.} \bibinfo{year}{2014}\natexlab{}.
\newblock \showarticletitle{Good debt or bad debt: Detecting semantic
  orientations in economic texts}.
\newblock \bibinfo{journal}{\emph{Journal of the Association for Information
  Science and Technology}} \bibinfo{volume}{65}, \bibinfo{number}{4}
  (\bibinfo{year}{2014}), \bibinfo{pages}{782--796}.
\newblock


\bibitem[OpenAI(2023)]%
        {openai2023gpt4}
\bibfield{author}{\bibinfo{person}{OpenAI}.} \bibinfo{year}{2023}\natexlab{}.
\newblock \bibinfo{title}{GPT-4 Technical Report}.
\newblock
\newblock
\showeprint[arxiv]{2303.08774}~[cs.CL]


\bibitem[Ouyang et~al\mbox{.}(2022)]%
        {ouyang2022training}
\bibfield{author}{\bibinfo{person}{Long Ouyang}, \bibinfo{person}{Jeffrey Wu},
  \bibinfo{person}{Xu Jiang}, \bibinfo{person}{Diogo Almeida},
  \bibinfo{person}{Carroll Wainwright}, \bibinfo{person}{Pamela Mishkin},
  \bibinfo{person}{Chong Zhang}, \bibinfo{person}{Sandhini Agarwal},
  \bibinfo{person}{Katarina Slama}, \bibinfo{person}{Alex Ray},
  {et~al\mbox{.}}} \bibinfo{year}{2022}\natexlab{}.
\newblock \showarticletitle{Training language models to follow instructions
  with human feedback}.
\newblock \bibinfo{journal}{\emph{Advances in Neural Information Processing
  Systems}}  \bibinfo{volume}{35} (\bibinfo{year}{2022}),
  \bibinfo{pages}{27730--27744}.
\newblock


\bibitem[Sanh et~al\mbox{.}(2022)]%
        {sanh2022multitask}
\bibfield{author}{\bibinfo{person}{Victor Sanh}, \bibinfo{person}{Albert
  Webson}, \bibinfo{person}{Colin Raffel}, \bibinfo{person}{Stephen~H Bach},
  \bibinfo{person}{Lintang Sutawika}, \bibinfo{person}{Zaid Alyafeai},
  \bibinfo{person}{Antoine Chaffin}, \bibinfo{person}{Arnaud Stiegler},
  \bibinfo{person}{Teven Le~Scao}, \bibinfo{person}{Arun Raja},
  {et~al\mbox{.}}} \bibinfo{year}{2022}\natexlab{}.
\newblock \showarticletitle{Multitask Prompted Training Enables Zero-Shot Task
  Generalization}. In \bibinfo{booktitle}{\emph{ICLR 2022-Tenth International
  Conference on Learning Representations}}.
\newblock


\bibitem[Scao et~al\mbox{.}(2022)]%
        {scao2022bloom}
\bibfield{author}{\bibinfo{person}{Teven~Le Scao}, \bibinfo{person}{Angela
  Fan}, \bibinfo{person}{Christopher Akiki}, \bibinfo{person}{Ellie Pavlick},
  \bibinfo{person}{Suzana Ili{\'c}}, \bibinfo{person}{Daniel Hesslow},
  \bibinfo{person}{Roman Castagn{\'e}}, \bibinfo{person}{Alexandra~Sasha
  Luccioni}, \bibinfo{person}{Fran{\c{c}}ois Yvon}, \bibinfo{person}{Matthias
  Gall{\'e}}, {et~al\mbox{.}}} \bibinfo{year}{2022}\natexlab{}.
\newblock \showarticletitle{Bloom: A 176b-parameter open-access multilingual
  language model}.
\newblock \bibinfo{journal}{\emph{arXiv preprint arXiv:2211.05100}}
  (\bibinfo{year}{2022}).
\newblock


\bibitem[Shah et~al\mbox{.}(2022)]%
        {shah2022flue}
\bibfield{author}{\bibinfo{person}{Raj~Sanjay Shah}, \bibinfo{person}{Kunal
  Chawla}, \bibinfo{person}{Dheeraj Eidnani}, \bibinfo{person}{Agam Shah},
  \bibinfo{person}{Wendi Du}, \bibinfo{person}{Sudheer Chava},
  \bibinfo{person}{Natraj Raman}, \bibinfo{person}{Charese Smiley},
  \bibinfo{person}{Jiaao Chen}, {and} \bibinfo{person}{Diyi Yang}.}
  \bibinfo{year}{2022}\natexlab{}.
\newblock \showarticletitle{WHEN FLUE MEETS FLANG: Benchmarks and Large
  Pre-trained Language Model for Financial Domain}.
\newblock \bibinfo{journal}{\emph{arXiv preprint arXiv:2211.00083}}
  (\bibinfo{year}{2022}).
\newblock


\bibitem[Sinha and Khandait(2021)]%
        {sinha2021impact}
\bibfield{author}{\bibinfo{person}{Ankur Sinha} {and} \bibinfo{person}{Tanmay
  Khandait}.} \bibinfo{year}{2021}\natexlab{}.
\newblock \showarticletitle{Impact of news on the commodity market: Dataset and
  results}. In \bibinfo{booktitle}{\emph{Advances in Information and
  Communication: Proceedings of the 2021 Future of Information and
  Communication Conference (FICC), Volume 2}}. Springer,
  \bibinfo{pages}{589--601}.
\newblock


\bibitem[Soun et~al\mbox{.}(2022)]%
        {soun2022accurate}
\bibfield{author}{\bibinfo{person}{Yejun Soun}, \bibinfo{person}{Jaemin Yoo},
  \bibinfo{person}{Minyong Cho}, \bibinfo{person}{Jihyeong Jeon}, {and}
  \bibinfo{person}{U Kang}.} \bibinfo{year}{2022}\natexlab{}.
\newblock \showarticletitle{Accurate Stock Movement Prediction with
  Self-supervised Learning from Sparse Noisy Tweets}. In
  \bibinfo{booktitle}{\emph{2022 IEEE International Conference on Big Data (Big
  Data)}}. IEEE, \bibinfo{pages}{1691--1700}.
\newblock


\bibitem[Taori et~al\mbox{.}(2023)]%
        {alpaca}
\bibfield{author}{\bibinfo{person}{Rohan Taori}, \bibinfo{person}{Ishaan
  Gulrajani}, \bibinfo{person}{Tianyi Zhang}, \bibinfo{person}{Yann Dubois},
  \bibinfo{person}{Xuechen Li}, \bibinfo{person}{Carlos Guestrin},
  \bibinfo{person}{Percy Liang}, {and} \bibinfo{person}{Tatsunori~B.
  Hashimoto}.} \bibinfo{year}{2023}\natexlab{}.
\newblock \bibinfo{title}{Stanford Alpaca: An Instruction-following LLaMA
  model}.
\newblock
  \bibinfo{howpublished}{\url{https://github.com/tatsu-lab/stanford_alpaca}}.
\newblock


\bibitem[Touvron et~al\mbox{.}(2023)]%
        {touvron2023llama}
\bibfield{author}{\bibinfo{person}{Hugo Touvron}, \bibinfo{person}{Thibaut
  Lavril}, \bibinfo{person}{Gautier Izacard}, \bibinfo{person}{Xavier
  Martinet}, \bibinfo{person}{Marie-Anne Lachaux},
  \bibinfo{person}{Timoth{\'e}e Lacroix}, \bibinfo{person}{Baptiste
  Rozi{\`e}re}, \bibinfo{person}{Naman Goyal}, \bibinfo{person}{Eric Hambro},
  \bibinfo{person}{Faisal Azhar}, {et~al\mbox{.}}}
  \bibinfo{year}{2023}\natexlab{}.
\newblock \showarticletitle{Llama: Open and efficient foundation language
  models}.
\newblock \bibinfo{journal}{\emph{arXiv preprint arXiv:2302.13971}}
  (\bibinfo{year}{2023}).
\newblock


\bibitem[Wang et~al\mbox{.}(2022)]%
        {wang2022self}
\bibfield{author}{\bibinfo{person}{Yizhong Wang}, \bibinfo{person}{Yeganeh
  Kordi}, \bibinfo{person}{Swaroop Mishra}, \bibinfo{person}{Alisa Liu},
  \bibinfo{person}{Noah~A Smith}, \bibinfo{person}{Daniel Khashabi}, {and}
  \bibinfo{person}{Hannaneh Hajishirzi}.} \bibinfo{year}{2022}\natexlab{}.
\newblock \showarticletitle{Self-Instruct: Aligning Language Model with Self
  Generated Instructions}.
\newblock \bibinfo{journal}{\emph{arXiv preprint arXiv:2212.10560}}
  (\bibinfo{year}{2022}).
\newblock


\bibitem[Wei et~al\mbox{.}(2021)]%
        {wei2021finetuned}
\bibfield{author}{\bibinfo{person}{Jason Wei}, \bibinfo{person}{Maarten Bosma},
  \bibinfo{person}{Vincent~Y Zhao}, \bibinfo{person}{Kelvin Guu},
  \bibinfo{person}{Adams~Wei Yu}, \bibinfo{person}{Brian Lester},
  \bibinfo{person}{Nan Du}, \bibinfo{person}{Andrew~M Dai}, {and}
  \bibinfo{person}{Quoc~V Le}.} \bibinfo{year}{2021}\natexlab{}.
\newblock \showarticletitle{Finetuned language models are zero-shot learners}.
\newblock \bibinfo{journal}{\emph{arXiv preprint arXiv:2109.01652}}
  (\bibinfo{year}{2021}).
\newblock


\bibitem[Wu et~al\mbox{.}(2018)]%
        {wu2018hybrid}
\bibfield{author}{\bibinfo{person}{Huizhe Wu}, \bibinfo{person}{Wei Zhang},
  \bibinfo{person}{Weiwei Shen}, {and} \bibinfo{person}{Jun Wang}.}
  \bibinfo{year}{2018}\natexlab{}.
\newblock \showarticletitle{Hybrid deep sequential modeling for social
  text-driven stock prediction}. In \bibinfo{booktitle}{\emph{Proceedings of
  the 27th ACM international conference on information and knowledge
  management}}. \bibinfo{pages}{1627--1630}.
\newblock


\bibitem[Wu et~al\mbox{.}(2023)]%
        {wu2023bloomberggpt}
\bibfield{author}{\bibinfo{person}{Shijie Wu}, \bibinfo{person}{Ozan Irsoy},
  \bibinfo{person}{Steven Lu}, \bibinfo{person}{Vadim Dabravolski},
  \bibinfo{person}{Mark Dredze}, \bibinfo{person}{Sebastian Gehrmann},
  \bibinfo{person}{Prabhanjan Kambadur}, \bibinfo{person}{David Rosenberg},
  {and} \bibinfo{person}{Gideon Mann}.} \bibinfo{year}{2023}\natexlab{}.
\newblock \showarticletitle{Bloomberggpt: A large language model for finance}.
\newblock \bibinfo{journal}{\emph{arXiv preprint arXiv:2303.17564}}
  (\bibinfo{year}{2023}).
\newblock


\bibitem[Xie et~al\mbox{.}(2023)]%
        {xie2023wall}
\bibfield{author}{\bibinfo{person}{Qianqian Xie}, \bibinfo{person}{Weiguang
  Han}, \bibinfo{person}{Yanzhao Lai}, \bibinfo{person}{Min Peng}, {and}
  \bibinfo{person}{Jimin Huang}.} \bibinfo{year}{2023}\natexlab{}.
\newblock \showarticletitle{The Wall Street Neophyte: A Zero-Shot Analysis of
  ChatGPT Over MultiModal Stock Movement Prediction Challenges}.
\newblock \bibinfo{journal}{\emph{arXiv preprint arXiv:2304.05351}}
  (\bibinfo{year}{2023}).
\newblock


\bibitem[Xu and Cohen(2018)]%
        {xu2018stock}
\bibfield{author}{\bibinfo{person}{Yumo Xu} {and} \bibinfo{person}{Shay~B
  Cohen}.} \bibinfo{year}{2018}\natexlab{}.
\newblock \showarticletitle{Stock movement prediction from tweets and
  historical prices}. In \bibinfo{booktitle}{\emph{Proceedings of the 56th
  Annual Meeting of the Association for Computational Linguistics (Volume 1:
  Long Papers)}}. \bibinfo{pages}{1970--1979}.
\newblock


\bibitem[Yang et~al\mbox{.}(2020)]%
        {yang2020finbert}
\bibfield{author}{\bibinfo{person}{Yi Yang}, \bibinfo{person}{Mark
  Christopher~Siy Uy}, {and} \bibinfo{person}{Allen Huang}.}
  \bibinfo{year}{2020}\natexlab{}.
\newblock \showarticletitle{Finbert: A pretrained language model for financial
  communications}.
\newblock \bibinfo{journal}{\emph{arXiv preprint arXiv:2006.08097}}
  (\bibinfo{year}{2020}).
\newblock


\bibitem[Zhang et~al\mbox{.}(2022)]%
        {zhang2022opt}
\bibfield{author}{\bibinfo{person}{Susan Zhang}, \bibinfo{person}{Stephen
  Roller}, \bibinfo{person}{Naman Goyal}, \bibinfo{person}{Mikel Artetxe},
  \bibinfo{person}{Moya Chen}, \bibinfo{person}{Shuohui Chen},
  \bibinfo{person}{Christopher Dewan}, \bibinfo{person}{Mona Diab},
  \bibinfo{person}{Xian Li}, \bibinfo{person}{Xi~Victoria Lin},
  {et~al\mbox{.}}} \bibinfo{year}{2022}\natexlab{}.
\newblock \showarticletitle{Opt: Open pre-trained transformer language models}.
\newblock \bibinfo{journal}{\emph{arXiv preprint arXiv:2205.01068}}
  (\bibinfo{year}{2022}).
\newblock


\bibitem[Zhang et~al\mbox{.}(2021)]%
        {zhang2021mengzi}
\bibfield{author}{\bibinfo{person}{Zhuosheng Zhang}, \bibinfo{person}{Hanqing
  Zhang}, \bibinfo{person}{Keming Chen}, \bibinfo{person}{Yuhang Guo},
  \bibinfo{person}{Jingyun Hua}, \bibinfo{person}{Yulong Wang}, {and}
  \bibinfo{person}{Ming Zhou}.} \bibinfo{year}{2021}\natexlab{}.
\newblock \showarticletitle{Mengzi: Towards lightweight yet ingenious
  pre-trained models for chinese}.
\newblock \bibinfo{journal}{\emph{arXiv preprint arXiv:2110.06696}}
  (\bibinfo{year}{2021}).
\newblock


\end{thebibliography}

\end{document}